\documentclass[runningheads]{llncs}

\usepackage{eccv}

\usepackage{eccvabbrv}

\usepackage{graphicx}
\usepackage{booktabs}

\usepackage[accsupp]{axessibility}  

\usepackage{hyperref}

\usepackage{orcidlink}
\usepackage{float}
\usepackage{xcolor}
\usepackage[normalem]{ulem}
\usepackage{comment}
\usepackage{multirow}

\newcommand\AP[1]{\textcolor{black}{#1}}

\usepackage{appendix}

\usepackage{enumitem}
\usepackage{listings}
\usepackage{minted}
\usepackage[linesnumbered,ruled,vlined]{algorithm2e}
\usepackage[T1]{fontenc}

\begin{document}
\title{A Benchmark for Semi-supervised Multi-modal Crowd Counting} 

\titlerunning{A Benchmark for Semi-supervised Multi-modal Crowd Counting}

\author{Haoliang Meng\inst{1} \and
Xiaopeng Hong\inst{1,2}\thanks{Corresponding author.} \and
Yabin Wang\inst{1} \and
Wangmeng Zuo\inst{1,2}}

\authorrunning{H. Meng et al.}

\institute{Harbin Institute of Technology \and
Pengcheng Laboratory\\
\email{menghaoliang2002@163.com}\quad  \email{hongxiaopeng@ieee.org}\quad \\ \email{wangyabin@outlook.com} \quad \email{cswmzuo@gmail.com}}

\maketitle

\begin{abstract}
    This paper constructs the first benchmark on semi-supervised multi-modal crowd counting. To lay the foundation for this unexplored task, we first formulate the semi-supervised multi-modal setting and a standardized protocol that specifies the labeled-unlabeled data partition across different labeled ratios. Next, to establish solid reference points, we carefully tailor a diverse set of representative baselines, including existing fully supervised multi-modal methods and semi-supervised single-modal methods. Then, we carefully evaluate their performance under our proposed benchmark.
    Codes and the data partition will be released on \url{https://github.com/HenryCilence/Semi-supervised-Multimodal-Crowd-Counting}.
\end{abstract}

\section{Introduction}
Multi-modal crowd counting~\cite{meng2024multi, meng2025free, liu2021cross, peng2020rgb} aims to estimate crowd density and distribution by integrating heterogeneous sensing data, such as RGB, thermal, and depth. By leveraging complementary cues, it significantly enhances accuracy and robustness, especially in challenging conditions where single-modal methods are unreliable. However, existing methods are predominantly developed under fully supervised settings, which heavily depend on large-scale annotated data. 

While semi-supervised learning has been explored in RGB-based counting to alleviate annotation dependency~\cite{lin2022semi,lin2025semi}, single-modal methods remain highly vulnerable to extreme illuminations or severe occlusions. Under such conditions, the generated pseudo-labels or predictions are often extremely noisy, leading to catastrophic confirmation bias~\cite{arazo2020pseudo} during semi-supervised training. Fortunately, heterogeneous sensors (\textit{e.g.}, optical-thermal cameras) are typically co-registered in hardware~\cite{liu2021cross,peng2020rgb}, which suggests that massive amounts of \textit{unannotated multi-modal image pairs} can be acquired efficiently. Thus, extending semi-supervised learning to the multi-modal domain is not only crucial for breaking the bottleneck of single-modal methods but also highly feasible in practice.

In this paper, we pioneer the task of semi-supervised multi-modal crowd counting. To lay the foundation for this unexplored task, we first formulate a standardized protocol that specifies the labeled-unlabeled data partition across different labeled-unlabeled ratios. Next, to establish solid reference points, we carefully tailor a diverse set of representative baselines, including existing fully supervised multi-modal methods and semi-supervised single-modal methods, thereby ultimately constructing a comprehensive benchmark.

\section{Related Work}
\subsection{Fully supervised multi-modal crowd counting}
To overcome the limitations of single-modal crowd counting~\cite{ma2019bayesian,ma2020learning,ma2021learning,lin2022boosting,lin2024gramformer,lin2021direct,shang20262d}, recent studies explore multi-modal approaches that fuse information from heterogeneous sources such as thermal imagery~\cite{meng2024multi,liu2021cross}, depth~\cite{lian2019density,lian2021locating}, and LiDAR~\cite{lesani2020development,chen2016svm}. Most existing works therefore focus on designing effective modality fusion or alignment modules. Representative strategies include cross-modal interaction through cross-modal attention~\cite{wu2022multimodal}, pixel-level cross-modal adversary‌~\cite{peng2020rgb}, asymmetric modal fusion~\cite{wang2025asymmetric}, decision-level density fusion~\cite{cheng2024late}, multi-modal information decomposition~\cite{mu2025misf}, and multi-scale modal deformation~\cite{liu2023rgb}. Other researchers propose intermediate branches that gradually converge information from two source modals through gated modal collaborative~\cite{liu2021cross}, auxiliary broker modality~\cite{meng2024multi}, and modal difference composition~\cite{zhou2023mc}.

However, these fusion methods rely on accurate annotations for optimization, which are limited under the semi-supervised setting. It is an urgent demand to tailor these methods to the semi-supervised setting and further develop semi-supervised multi-modal models.

\subsection{Semi-supervised RGB crowd counting}
Due to the high cost of dense point annotations, semi-supervised crowd counting has gained increasing attention. Representative approaches employ dual models based on Mean Teacher~\cite{tarvainen2017mean}, and organize self-supervision signals by enforcing their prediction consistency to make efficient use of unlabeled data. For instance, OT-M~\cite{lin2023optimal} enforces consistency between Optimal Transport-generated point maps and student predictions. SUA~\cite{meng2021spatial} enforces the consistency between dual models on main and several pretext tasks. There are also other methods that employ independent density decoders for prediction consistency. For instance, GP~\cite{sindagi2020learning} introduces a Gaussian Process-based iteration as the counterpart of the decoding network and minimizes the distance between their outputs. P$^3$Net~\cite{lin2025semi} enhances consistency between two independent decoders with interleaved density scales. There are also other semi-supervised paradigms based on active learning strategies~\cite{zhao2020active} and foreground-background relationships~\cite{lin2022semi}. 

\section{Benchmark and Baseline Construction}
\subsection{Semi-supervised setting}
Given that RGB and thermal images are naturally co-registered, and the annotation bottleneck primarily lies in point labeling, we assume that both modalities are available for all training samples, while only a small subset is annotated. Following prior RGB-based semi-supervised settings~\cite{lin2022semi,lin2025semi}, we adopt labeling ratios of 5\%, 10\%, and 40\%, corresponding to stringent, moderate, and relaxed annotation budgets, respectively. For each ratio, a fixed subset of training samples is selected as labeled data, while the remaining samples are treated as unlabeled. Formally, given the training set containing $N$ samples and the labeling ratio $\mu$, we split it into a labeled subset $\mathcal{D}_l=\{(x_l^v,x_l^t,y_l)\}_{l=1}^{\mu N}$ and an unlabeled subset $\mathcal{D}_u=\{(x_u^v,x_u^t)\}_{u=1}^{(1-\mu)N}$. To avoid sampling bias caused by the acquisition order of images, the labeled subset is constructed via fixed-interval sampling over the filename-sorted training set. All labeled splits will be released upon acceptance of the paper.

Model performances are evaluated on two widely used RGB-Thermal crowd counting datasets, RGBT-CC~\cite{liu2021cross} and DroneRGBT~\cite{peng2020rgb}. Considering that the two datasets contain thermal images acquired under different sensing modes, all models are trained and evaluated separately on each dataset rather than merging their training sets. The numbers of training and testing samples under different labeling ratios are summarized in \cref{tab:split}. Following prior multi-modal crowd counting works, we adopt GAME~\cite{guerrero2015extremely} and RMSE~\cite{sindagi2019multi} as the evaluation metrics. Compared with evaluating only the total count, GAME imposes stricter constraints on regional density estimation.

\begin{equation}
  \mathrm{GAME}(l) = \frac{1}{N}\sum_{i=1}^{N}\sum_{j=1}^{4^l} \left|\hat{\mathcal{D}_i^j}-\mathcal{D}_i^j\right|,
\end{equation}
\begin{equation}
  \mathrm{RMSE} = \left(\frac{1}{N}\sum_{i=1}^{N} (\hat{\mathcal{D}_i}-\mathcal{D}_i)^2\right)^{\frac{1}{2}},
\end{equation}
where $N$ denotes the number of testing samples, $\hat{\mathcal{D}_i}$ and ${\mathcal{D}_i}$ denote the estimated count and the ground-truth count for the $i$-th sample, $\hat{\mathcal{D}_i^j}$ and ${\mathcal{D}_i^j}$ denote the estimated count and the ground-truth count for the $j$-th region of the $i$-th image pair, $4^l$ denotes the number of the divided non-overlapping regions and $l\in\{0,1,2,3\}$.

\begin{table}[!h]
    \centering
    \caption{The numbers of training and testing samples under different labeling ratios of RGBT-CC and DroneRGBT.}
    \label{tab:split}
    \small
    \begin{tabular}{c | c c | c c | c c | c}
        \toprule
        & \multicolumn{6}{c|}{Labeling ratios of training samples} & \\
        Dataset & \multicolumn{2}{c|}{5\%} & \multicolumn{2}{c|}{10\%} & \multicolumn{2}{c|}{40\%} & Testing \\
        & Labeled & Unlabeled & Labeled & Unlabeled & Labeled & Unlabeled & \\
        \midrule
        RGBT-CC     &  51 & 979 & 103 & 927 & 412 & 618 & 800 \\
        DroneRGBT   &  90 & 1710 & 180 & 1620 & 722 & 1078 & 1807 \\
        \bottomrule
    \end{tabular}
\end{table}

\subsection{Baseline construction}
To thoroughly investigate the proposed semi-supervised multi-modal problem, we carefully select a diverse set of baseline methods and adapt them for evaluation on our constructed benchmark. The baselines are derived from two categories of representative methods: fully supervised multi-modal approaches and semi-supervised RGB-based approaches. The description for the selected baseline methods is summarized in \cref{tab:baseline}.

For fully supervised multi-modal methods, we select representative open-source counting models spanning both dual-branch interaction and multi-branch fusion paradigms. For each approach, we evaluate its original fully supervised version using only labeled samples, and its semi-supervised extension built upon the Mean Teacher framework~\cite{tarvainen2017mean}. Specifically, the compared models include: Dual-branch BL Baseline~\cite{ma2019bayesian}, IADM~\cite{liu2021cross}, CAGNet~\cite{yang2024cagnet}, DEFNet~\cite{zhou2022defnet}, MC$^3$Net~\cite{zhou2023mc}, BM~\cite{meng2024multi}.

For semi-supervised RGB-based methods, we select four representative open-source methods from the past few years, including the pseudo-labeling-based method MTCP~\cite{zhu2023multi}, the consistency-based method IRAST~\cite{liu2020semi}, the correlation-based method DACount~\cite{lin2022semi}, and the recent state-of-the-art method P$^3$Net~\cite{lin2025semi}. To adapt these RGB-only semi-supervised approaches to the multi-modal setting, we feed RGB-Thermal pairs through their encoders and \AP{integrate their representations via a learnable gated fusion module} to build a dual-branch variant for comparison. 

\begin{table}[!h]
    \caption{Description for the selected baseline methods in our proposed benchmark. The top contains fully supervised multi-modal methods, and the bottom contains semi-supervised RGB-based methods.}
    \centering
    \small
    \begin{tabular}{ccccc}
    \toprule
        Method & Venue & Input data & Backbone & Model Arch.\\
        \midrule
        Dual-branch BL~\cite{ma2019bayesian} & ICCV 2019 & RGB-T & VGG-19 & Dual-branch\\
        IADM~\cite{liu2021cross} & CVPR 2021 & RGB-T & VGG-19 & Dual-branch\\
        DEFNet~\cite{zhou2022defnet} & TITS 2022 & RGB-T & VGG-16 & Dual-branch\\
        MC$^3$Net~\cite{zhou2023mc} & TITS 2023 & RGB-T & ConvNeXt-S & Triple-branch\\
        CAGNet~\cite{yang2024cagnet} & ESWA 2024 & RGB-T & VGG-16 & Dual-branch\\
        BM~\cite{meng2024multi} & ECCV 2024 & RGB-T & VGG-19-Trans. & Triple-branch\\
        \midrule
        IRAST~\cite{liu2020semi} & ECCV 2020 & RGB & VGG-16 & Multi-branch\\
        DACount~\cite{lin2022semi} & ACM MM 2022 & RGB & VGG-19 & Dual-branch\\
        MTCP~\cite{zhu2023multi} & TNNLS 2024 & RGB & VGG-16 & Triple-branch\\
        P$^3$Net~\cite{lin2025semi} & TPAMI 2025 & RGB & VGG-19 & Dual-branch\\    
    \bottomrule
    \end{tabular}
    \label{tab:baseline}
\end{table}

These selected baselines provide comprehensive coverage of mainstream counting methods. The adaptation paradigms of two categories of methods are described in algorithms~\ref{alg:fully} and ~\ref{alg:semi}.

\begin{algorithm}[!h]
\small
\SetAlgoLined
\caption{Semi-supervised extension of fully supervised multi-modal methods}
\label{alg:fully}
\KwIn{Fully supervised multi-modal model $f$ with pretrained weight $\theta$ and its original supervised regression loss $\mathcal{L}_{sup}$, labeled training set $\mathcal{D}_l=\{(x_l^v,x_l^t,y_l)\}$, unlabeled training set $\mathcal{D}_u=\{(x_u^v,x_u^t)\}$}
\KwOut{Optimized extended model under our semi-supervised setting}
Instantiate student model $f_s$ with $f$ and initialize its parameters $\theta_s \leftarrow \theta$\;
Instantiate teacher model $f_t$ with $f$ and initialize its parameters $\theta_t \leftarrow \theta$\;
\For{Each training iteration}{
Sample labeled batches $(x_l^v,x_l^t,y_l)\sim\mathcal{D}_l$ and unlabeled batches $(x_u^v,x_u^t)\sim\mathcal{D}_u$\;
Obtain supervised prediction on labeled data: $\hat{y}_l = f_s(x_l^v,x_l^t)$\;
Compute supervised density regression loss: $\mathcal{L}_{sup}(\hat{y}_l,y_l)$\;
Generate augmented inputs for unlabeled data: $(\tilde{x}_u^v,\tilde{x}_u^t)$ and $(\hat{x}_u^v,\hat{x}_u^t)$ via color jitter, flipping or Gaussian blur\;
Obtain teacher and student prediction on unlabeled data: $\bar{y}_u = f_t(\tilde{x}_u^v,\tilde{x}_u^t),\quad \hat{y}_u = f_s(\hat{x}_u^v,\hat{x}_u^t)$\;
Compute consistency loss: $\mathcal{L}_{con}(\hat{y}_u,\bar{y}_u)$\;
Compute total loss: $\mathcal{L} = \mathcal{L}_{sup}(\hat{y}_l,y_l) + \lambda \mathcal{L}_{con}(\hat{y}_u,\bar{y}_u)$\;
Update student parameters $\theta_s$ using $\mathcal{L}$\;
Update teacher parameters using EMA: $\theta_t \leftarrow \alpha \theta_t + (1-\alpha)\theta_s$\;
}
Return optimized student model $f_s$ with parameters $\theta_s$.
\end{algorithm}

\begin{algorithm}[!h]
\small
\SetAlgoLined
\caption{Semi-supervised extension of semi-supervised RGB-based methods}
\label{alg:semi}
\KwIn{Semi-supervised RGB-based model $f$, the original supervised and unsupervised loss $\mathcal L_{s}$ and $\mathcal L_{us}$, labeled training set $\mathcal{D}_l=\{(x_l^v,x_l^t,y_l)\}$, unlabeled training set $\mathcal{D}_u=\{(x_u^v,x_u^t)\}$}
\KwOut{Optimized multi-modal variant under our semi-supervised setting}
Decompose the original RGB-based model $f$ into an encoder $E$ and a decoder (or remaining prediction module) $D$, \textit{i.e.}, $f(x)=D(E(x))$\;
Instantiate visual and thermal encoders with the same architecture and initialization as $E$: $E_v \leftarrow E, E_t \leftarrow E$, respectively\;
\For{Each training iteration}{
Sample labeled batches $(x_l^v,x_l^t,y_l)\sim\mathcal{D}_l$ and unlabeled batches $(x_u^v,x_u^t)\sim\mathcal{D}_u$\;
Extract visual and thermal features:
\[F_l^v = E_v(x_l^v), \quad F_l^t = E_t(x_l^t)\]
\[F_u^v = E_v(x_u^v), \quad F_u^t = E_t(x_u^t);\]
Compute their gating weights:
\[g_l^v=\sigma(\phi_v(F_l^v)), \quad g_l^t=\sigma(\phi_t(F_l^t)),\]
\[g_u^v=\sigma(\phi_v(F_u^v)), \quad g_u^t=\sigma(\phi_t(F_u^t));\]
Fuse the features:
\[F_l = F_l^v \odot g_l^v + F_l^t \odot g_l^t,\quad F_u = F_u^v \odot g_u^v + F_u^t \odot g_u^t,\]
where $\phi_v(\cdot)$ and $\phi_t(\cdot)$ are learnable gating functions, $\sigma(\cdot)$ denotes the sigmoid activation, and $\odot$ denotes element-wise multiplication\;
Feed the fused features into the original decoder:
\[\hat{y}_l = D(F_l), \quad \hat{y}_u = D(F_u);\]
Compute the original supervised and unsupervised loss $\mathcal L_{s}$ and $\mathcal L_{us}$ using $\hat{y}_l$ and $\hat{y}_u$, respectively\;
Update all parameters using $\mathcal L_{s}$ and $\mathcal L_{us}$\;
}
Return the optimized multi-modal variant.
\end{algorithm}

\section{Experiments}
\subsection{Quantitative results}
We evaluate all methods under our constructed benchmark, and the results are listed in \cref{tab:result_rgbtcc_5} and \cref{tab:result_drone}. In each table, the top part contains fully supervised RGB-T crowd counting methods. We report their original performance using only labeled data and their semi-supervised variants equipped with Mean Teacher~\cite{tarvainen2017mean}. The bottom part contains semi-supervised RGB crowd counting methods, which are extended to dual-branch multi-modal versions for comparison. All results are reproduced by ourselves.

\subsection{Complexity comparison}
We also compare the computational complexity of different methods in terms of learnable parameters and floating-point operations (FLOPs), as shown in \cref{tab:result_complexity}.

\begin{table}[!h]
\small
  \centering
      \caption{Complexity comparison with the state-of-the-art RGB-T crowd counting methods. All models receive a 3-channel visual image and a 3-channel thermal image of size 640$\times$480 as input. $*$ denotes the result reproduced by ourselves.}
      \begin{tabular}{ccc|cc}
      \toprule
  Method & Encoder Arch. & Venue & Param. (M) & FLOPs (G)\\
  \midrule
  MIDD~\cite{tu2021multi} & VGG-16 & TIP 2021 & 53.31 & 605.32\\
  CGFNet~\cite{wang2021cgfnet} & VGG-16 & TCSVT 2021 & 66.38 & 855.80\\
  DEFNet~\cite{zhou2022defnet} & VGG-16 & TITS 2022 & 45.33 & 494.10\\
  MC$^3$Net* \cite{zhou2023mc} & ConvNeXt-S & TITS 2023 & 260.52 & 713.99\\
  CAGNet~\cite{yang2024cagnet} & VGG-16 & ESWA 2024 & 68.15 & 557.85\\
  BM*~\cite{meng2024multi} & VGG-19-Trans & ECCV 2024 & 40.55 & 529.55\\
  MISFNet~\cite{mu2025misf} & VGG-16 & TMM 2025 & 82.45 & 1144.43\\
  \bottomrule
      \end{tabular}
    \label{tab:result_complexity}
\end{table}

\subsection{Qualitative results}
\begin{figure}[!h]
    \centering
    \includegraphics[width=1\linewidth]{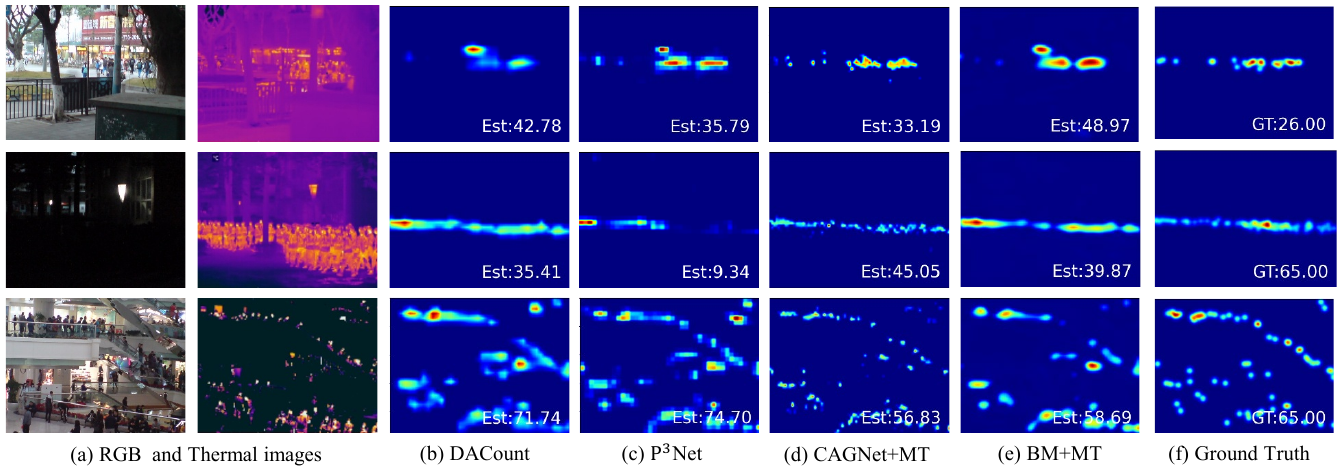}
    \caption{Density maps from the testing set of RGBT-CC~\cite{liu2021cross} generated by different methods. All models are trained under 5\% labeled training data.}
    \label{fig:visual}
\end{figure}

We visualize representative density maps on RGBT-CC~\cite{liu2021cross} predicted by representative baseline methods for further comparison, as shown in \cref{fig:visual}.

\begin{table}[!t]
\scriptsize
  \caption{Comparison of baseline methods on RGBT-CC~\cite{liu2021cross} with \textbf{5\%}, \textbf{10\%}, and \textbf{40\%} labeled training samples.}
  \centering
  \resizebox{0.75\textwidth}{!}{
  \begin{tabular}{c|ccccc}
  \toprule
  Method & GAME(0) & GAME(1) & GAME(2) & GAME(3) & RMSE \\
  \midrule
  \midrule
  \multicolumn{6}{c}{Labeled Percentage: 5\%}\\
  \midrule
  BL~\cite{ma2019bayesian} & 22.45 & 26.44 & 31.08 & 37.81 & 48.46\\
  BL+MT & 20.51 & 24.35 & 29.08 & 35.71 & 42.59\\
  IADM~\cite{liu2021cross} & 19.53 & 23.74 & 28.62 & 36.34 & 36.53\\
  IADM+MT & 18.57 & 22.92 & 27.93 & 35.71 & 35.90\\
  DEFNet~\cite{zhou2022defnet} & 30.17 & 34.12 & 39.14 & 45.54 & 66.76\\
  DEFNet+MT & 19.25 & 24.22 & 30.17 & 37.45 & 32.34\\ 
  MC$^3$Net~\cite{zhou2023mc} & 22.59 & 28.19 & 34.53 & 43.44 & 32.16\\  
  MC$^3$Net+MT & 15.96 & 20.09 & 24.75 & 32.40 & 29.77\\
  CAGNet~\cite{yang2024cagnet} & 20.46 & 24.72 & 29.77 & 36.82 & 38.71\\
  CAGNet+MT & 19.28 & 24.20 & 29.45 & 37.37 & 30.78\\
  BM~\cite{meng2024multi} & 20.12 & 24.96 & 30.15 & 37.29 & 42.87\\
  BM+MT & 17.65 & 22.55 & 28.11 & 35.36 & 35.23\\ 
  \midrule
  IRAST~\cite{liu2020semi} & 19.26 & 22.84 & 27.11 & 35.53 & 38.70\\
  DACount~\cite{lin2022semi} & 19.11 & 22.48 & 26.44 & 33.23 & 40.25\\
  MTCP~\cite{zhu2023multi} & 19.08 & 22.92 & 26.37 & 34.60 & 39.14\\
  P$^3$Net~\cite{lin2025semi} & 18.64 & 22.80 & 27.74 & 35.35 & 39.52\\
  \midrule
  \midrule
  \multicolumn{6}{c}{Labeled Percentage: 10\%}\\
  \midrule
BL~\cite{ma2019bayesian} & 20.40 & 24.69 & 29.75 & 37.15 & 42.73\\
BL+MT & 17.99 & 23.11 & 29.07 & 36.98 & 35.40\\
IADM~\cite{liu2021cross} & 19.24 & 23.72 & 29.50 & 37.44 & 35.55\\
IADM+MT & 16.75 & 20.48 & 25.14 & 32.01 & 32.74\\
DEFNet~\cite{zhou2022defnet} & 23.76 & 28.00 & 32.89 & 39.31 & 41.35\\
DEFNet+MT & 17.27 & 20.56 & 24.03 & 29.74 & 29.50\\ 
MC$^3$Net~\cite{zhou2023mc} & 20.60 & 30.21 & 38.91 & 46.62 & 36.16\\  
MC$^3$Net+MT & 14.90 & 19.71 & 24.97 & 34.09 & 22.18\\
CAGNet~\cite{yang2024cagnet} & 18.71 & 23.95 & 28.62 & 35.33 & 34.96\\
CAGNet+MT & 15.44 & 19.85 & 25.15 & 32.32 & 24.84\\
BM~\cite{meng2024multi} & 18.50 & 22.44 & 27.23 & 34.23 & 37.44\\ 
BM+MT & 17.78 & 21.71 & 26.14 & 32.71 & 36.03\\ 
\midrule
IRAST~\cite{liu2020semi} & 18.88 & 21.19 & 26.42 & 30.03 & 32.76\\
DACount~\cite{lin2022semi} & 15.36 & 18.92 & 23.07 & 29.97 & 29.82\\
MTCP~\cite{zhu2023multi} & 17.55 & 21.30 & 25.67 & 32.08 & 31.91\\
P$^3$Net~\cite{lin2025semi} & 17.88 & 22.24 & 26.97 & 34.64 & 38.13\\
  \midrule
  \midrule
  \multicolumn{6}{c}{Labeled Percentage: 40\%}\\
  \midrule
BL~\cite{ma2019bayesian} & 19.05 & 22.97 & 27.90 & 34.76 & 40.86\\
BL+MT & 17.00 & 20.51 & 24.51 & 30.96 & 35.97\\
IADM~\cite{liu2021cross} & 16.99 & 21.99 & 27.53 & 35.84 & 30.77\\
IADM+MT & 16.55 & 20.21 & 24.27 & 30.94 & 32.30\\
DEFNet~\cite{zhou2022defnet} & 18.74 & 23.77 & 28.96 & 35.63 & 31.01\\
DEFNet+MT & 16.65 & 19.52 & 22.77 & 28.74 & 30.22\\ 
MC$^3$Net~\cite{zhou2023mc} & 15.05 & 18.93 & 23.20 & 30.74 & 27.41\\  
MC$^3$Net+MT & 13.77 & 18.55 & 24.02 & 32.84 & 24.91\\
CAGNet~\cite{yang2024cagnet} & 17.18 & 20.96 & 24.72 & 31.12 & 26.62\\
CAGNet+MT & 16.07 & 19.79 & 24.50 & 30.95 & 26.68\\
BM~\cite{meng2024multi} & 15.80 & 19.54 & 24.07 & 30.94 & 29.07\\
BM+MT & 15.29 & 18.85 & 23.09 & 29.62 & 31.70\\ 
\midrule
IRAST~\cite{liu2020semi} & 13.22 & 16.49 & 21.00 & 28.81 & 22.69\\
DACount~\cite{lin2022semi} & 12.88 & 16.46 & 20.40 & 27.62 & 20.73\\
MTCP~\cite{zhu2023multi} & 12.73 & 17.15 & 21.94 & 28.28 & 23.50\\
P$^3$Net~\cite{lin2025semi} & 12.53 & 16.11 & 20.52 & 27.95 & 23.50\\
  
  \bottomrule
  \end{tabular}
  }
  \label{tab:result_rgbtcc_5}
\end{table}

\begin{table}[!t]
\scriptsize
  \caption{Comparison of baseline methods on DroneRGBT~\cite{peng2020rgb} with \textbf{5\%}, \textbf{10\%}, and \textbf{40\%} labeled training samples.}
  \centering
  \resizebox{0.75\textwidth}{!}{
  \begin{tabular}{c|ccccc}
  \toprule
  Method & GAME(0) & GAME(1) & GAME(2) & GAME(3) & RMSE \\
  \midrule
  \midrule
  \multicolumn{6}{c}{Labeled Percentage: 5\%}\\
  \midrule
  BL~\cite{ma2019bayesian} & 11.38 & 14.36 & 18.49 & 23.67 & 17.93\\
  BL+MT & 9.42 & 12.08 & 15.96 & 20.33 & 14.66\\
  IADM~\cite{liu2021cross} & 10.00 & 12.28 & 15.66 & 20.40 & 16.06\\
  IADM+MT & 9.46 & 12.00 & 15.30 & 19.39 & 15.71\\
  DEFNet~\cite{zhou2022defnet} & 11.07 & 13.21 & 15.94 & 19.27 & 18.51\\
  DEFNet+MT & 9.88 & 11.89 & 14.83 & 18.82 & 15.36\\ 
  MC$^3$Net~\cite{zhou2023mc} & 14.69 & 20.77 & 29.95 & 36.64 & 21.54\\  
  MC$^3$Net+MT & 9.40 & 11.57 & 14.62 & 18.63 & 14.48\\
  CAGNet~\cite{yang2024cagnet} & 9.82 & 12.12 & 15.01 & 18.86 & 16.14\\
  CAGNet+MT & 9.12 & 11.32 & 14.07 & 18.03 & 15.38\\
  BM~\cite{meng2024multi} & 9.73 & 11.93 & 15.04 & 19.35 & 14.93\\
  BM+MT & 9.62 & 11.30 & 14.48 & 18.47 & 15.58\\ 
  \midrule
  IRAST~\cite{liu2020semi} & 9.40 & 11.66 & 14.12 & 17.97 & 15.24\\
  DACount~\cite{lin2022semi} & 8.99 & 10.87 & 13.54 & 17.88 & 14.10\\
  MTCP~\cite{zhu2023multi} & 8.59 & 11.07 & 13.85 & 18.31 & 14.64\\
  P$^3$Net~\cite{lin2025semi} & 8.71 & 10.70 & 13.54 & 17.67 & 14.32\\
  \midrule
  \midrule
  \multicolumn{6}{c}{Labeled Percentage: 10\%}\\
  \midrule
BL~\cite{ma2019bayesian} & 9.04 & 11.61 & 14.99 & 19.30 & 14.55\\
BL+MT & 8.73 & 11.60 & 15.64 & 20.40 & 13.43\\
IADM~\cite{liu2021cross} & 8.77 & 11.34 & 14.70 & 19.04 & 14.42\\
IADM+MT & 8.21 & 10.54 & 13.63 & 17.69 & 12.97\\
DEFNet~\cite{zhou2022defnet} & 9.13 & 11.28 & 13.90 & 17.40 & 14.98\\
DEFNet+MT & 8.87 & 10.99 & 13.80 & 17.47 & 14.95\\ 
MC$^3$Net~\cite{zhou2023mc} & 10.55 & 12.82 & 16.33 & 20.50 & 15.35\\
MC$^3$Net+MT & 8.73 & 10.44 & 12.95 & 17.21 & 13.85\\
CAGNet~\cite{yang2024cagnet} & 9.55 & 11.97 & 14.87 & 18.91 & 15.90\\
CAGNet+MT & 8.44 & 10.32 & 13.02 & 16.87 & 13.61\\
BM~\cite{meng2024multi} & 9.47 & 11.94 & 15.10 & 19.22 & 15.65\\
BM+MT & 9.15 & 11.66 & 15.05 & 19.23 & 15.38\\ 
\midrule
IRAST~\cite{liu2020semi} & 9.18 & 10.62 & 13.93 & 17.79 & 15.02\\
DACount~\cite{lin2022semi} & 8.83 & 10.81 & 13.62 & 18.21 & 13.61\\
MTCP~\cite{zhu2023multi} & 8.90 & 11.04 & 13.57 & 17.21 & 14.48\\
P$^3$Net~\cite{lin2025semi} & 8.54 & 10.76 & 13.74 & 18.02 & 14.77\\
  \midrule
  \midrule
  \multicolumn{6}{c}{Labeled Percentage: 40\%}\\
  \midrule
BL~\cite{ma2019bayesian} & 8.80 & 10.71 & 13.45 & 17.36 & 14.46\\
BL+MT & 7.98 & 9.80 & 12.64 & 16.79 & 12.55\\
IADM~\cite{liu2021cross} & 8.18 & 10.26 & 13.12 & 17.07 & 13.04\\
IADM+MT & 7.62 & 9.77 & 12.74 & 16.87 & 11.95\\
DEFNet~\cite{zhou2022defnet} & 8.78 & 10.55 & 12.94 & 16.33 & 15.21\\
DEFNet+MT & 7.71 & 9.78 & 12.39 & 15.71 & 12.62\\ 
MC$^3$Net~\cite{zhou2023mc} & 8.57 & 10.56 & 13.85 & 18.13 & 13.05\\  
MC$^3$Net+MT & 7.73 & 9.79 & 13.29 & 17.97 & 11.89\\
CAGNet~\cite{yang2024cagnet} & 8.61 & 10.54 & 13.17 & 16.88 & 14.58\\
CAGNet+MT & 7.66 & 9.58 & 12.23 & 15.73 & 12.55\\
BM~\cite{meng2024multi} & 8.74 & 10.88 & 13.81 & 17.94 & 13.81\\
BM+MT & 7.62 & 9.84 & 12.78 & 16.74 & 12.43\\
\midrule
IRAST~\cite{liu2020semi} & 8.39 & 10.74 & 13.16 & 16.83 & 14.45\\
DACount~\cite{lin2022semi} & 7.17 & 9.20 & 12.10 & 16.73 & 11.35\\
MTCP~\cite{zhu2023multi} & 8.06 & 10.52 & 12.77 & 17.25 & 14.98\\
P$^3$Net~\cite{lin2025semi} & 8.77 & 11.34 & 14.70 & 19.04 & 14.42\\ 
  \bottomrule
  \end{tabular}
  }
  \label{tab:result_drone}
\end{table}


\section{Conclusion}
This paper establishes the first semi-supervised multi-modal crowd counting benchmark. To lay the foundation for this unexplored task, we first formulate a standardized protocol that specifies the labeled-unlabeled data partition across different labeled-unlabeled ratios. Next, to establish solid reference points, we carefully tailor a diverse set of representative baseline methods, including existing fully supervised multi-modal methods and semi-supervised single-modal methods. Finally, we carefully evaluate their performance under our proposed benchmark. We hope that our work opens new avenues for future studies of semi-supervised multi-modal crowd counting.

\section*{Acknowledgements}
This work was supported by the National Key R\&D Program of China (No. 2025YFC3811300) and the National Natural Science Foundation of China (Grant Nos. 62376070 and 62076195).

\clearpage

\bibliographystyle{splncs04}
\bibliography{egbib}

@String(ICME = {Int. Conf. Multimedia and Expo})

@String(AAAI = {AAAI})

@String(ICME  =	{ICME})

@inproceedings{meng2024multi,
  title={Multi-modal crowd counting via a broker modality},
  author={Meng, Haoliang and Hong, Xiaopeng and Wang, Chenhao and Shang, Miao and Zuo, Wangmeng},
  booktitle={European Conference on Computer Vision},
  pages={231--250},
  year={2024},
  organization={Springer}
}

@inproceedings{ma2019bayesian,
  title={Bayesian loss for crowd count estimation with point supervision},
  author={Ma, Zhiheng and Wei, Xing and Hong, Xiaopeng and Gong, Yihong},
  booktitle={Proceedings of the IEEE/CVF international conference on computer vision},
  pages={6142--6151},
  year={2019}
}

@inproceedings{liu2021cross,
  title={Cross-modal collaborative representation learning and a large-scale rgbt benchmark for crowd counting},
  author={Liu, Lingbo and Chen, Jiaqi and Wu, Hefeng and Li, Guanbin and Li, Chenglong and Lin, Liang},
  booktitle={Proceedings of the IEEE/CVF conference on computer vision and pattern recognition},
  pages={4823--4833},
  year={2021}
}

@article{zhou2023mc,
  title={MC$^3$Net: Multimodality cross-guided compensation coordination network for RGB-T crowd counting},
  author={Zhou, Wujie and Yang, Xun and Lei, Jingsheng and Yan, Weiqing and Yu, Lu},
  journal={IEEE Transactions on Intelligent Transportation Systems},
  volume={25},
  number={5},
  pages={4156--4165},
  year={2023},
  publisher={IEEE}
}

@article{tarvainen2017mean,
  title={Mean teachers are better role models: Weight-averaged consistency targets improve semi-supervised deep learning results},
  author={Tarvainen, Antti and Valpola, Harri},
  journal={Advances in neural information processing systems},
  volume={30},
  year={2017}
}

@article{lin2025semi,
  title={Semi-supervised counting via pixel-by-pixel density distribution modelling},
  author={Lin, Hui and Ma, Zhiheng and Ji, Rongrong and Wang, Yaowei and Su, Zhou and Hong, Xiaopeng and Meng, Deyu},
  journal={IEEE Transactions on Pattern Analysis and Machine Intelligence},
  year={2025},
  publisher={IEEE}
}

@article{zhou2022defnet,
  title={DEFNet: Dual-branch enhanced feature fusion network for RGB-T crowd counting},
  author={Zhou, Wujie and Pan, Yi and Lei, Jingsheng and Ye, Lv and Yu, Lu},
  journal={IEEE Transactions on Intelligent Transportation Systems},
  volume={23},
  number={12},
  pages={24540--24549},
  year={2022},
  publisher={IEEE}
}

@article{yang2024cagnet,
  title={CAGNet: Coordinated attention guidance network for RGB-T crowd counting},
  author={Yang, Xun and Zhou, Wujie and Yan, Weiqing and Qian, Xiaohong},
  journal={Expert Systems with Applications},
  volume={243},
  pages={122753},
  year={2024},
  publisher={Elsevier}
}

@article{cheng2024late,
  title={Late better than early: A decision-level information fusion approach for RGB-Thermal crowd counting with illumination awareness},
  author={Cheng, Jian and Feng, Chen and Xiao, Yang and Cao, Zhiguo},
  journal={Neurocomputing},
  volume={594},
  pages={127888},
  year={2024},
  publisher={Elsevier}
}

@article{mu2025misf,
  title={MISF-Net: Modality-Invariant and-Specific Fusion Network for RGB-T Crowd Counting},
  author={Mu, Baoyang and Shao, Feng and Xie, Zhenxuan and Chen, Hangwei and Zhu, Zhongjie and Li, Xiaoer and Jiang, Qiuping},
  journal={IEEE Transactions on Multimedia},
  year={2025},
  publisher={IEEE}
}

@article{liu2023rgb,
  title={RGB-T Multi-Modal Crowd Counting Based on Transformer},
  author={Liu, Zhengyi and Wu, Wei and Tan, Yacheng and Zhang, Guanghui},
  journal={The 33rd British Machine Vision Conference 2022},
  year={2022}
}

@inproceedings{peng2020rgb,
  title={Rgb-t crowd counting from drone: A benchmark and mmccn network},
  author={Peng, Tao and Li, Qing and Zhu, Pengfei},
  booktitle={Proceedings of the Asian conference on computer vision},
  year={2020}
}

@inproceedings{meng2025free,
  title={Free Lunch Enhancements for Multi-modal Crowd Counting},
  author={Meng, Haoliang and Hong, Xiaopeng and Lai, Zhengqin and Shang, Miao},
  booktitle={Proceedings of the Computer Vision and Pattern Recognition Conference},
  pages={14013--14023},
  year={2025}
}

@inproceedings{lin2022semi,
  title={Semi-supervised crowd counting via density agency},
  author={Lin, Hui and Ma, Zhiheng and Hong, Xiaopeng and Wang, Yaowei and Su, Zhou},
  booktitle={Proceedings of the 30th ACM International Conference on Multimedia},
  pages={1416--1426},
  year={2022}
}

@article{zhu2023multi,
  title={Multi-task credible pseudo-label learning for semi-supervised crowd counting},
  author={Zhu, Pengfei and Li, Jingqing and Cao, Bing and Hu, Qinghua},
  journal={IEEE Transactions on Neural Networks and Learning Systems},
  volume={35},
  number={8},
  pages={10394--10406},
  month={August},
  year={2024},
  publisher={IEEE}
}

@inproceedings{liu2020semi,
 title={Semi-Supervised Crowd Counting via Self-Training on Surrogate Tasks},
 author={Liu, Yan and Liu, Lingqiao and Wang, Peng and Zhang, Pingping and Lei, Yinjie},
 booktitle={European Conference on Computer Vision},
 year={2020}
}

@article{tu2021multi,
  title={Multi-interactive dual-decoder for RGB-thermal salient object detection},
  author={Tu, Zhengzheng and Li, Zhun and Li, Chenglong and Lang, Yang and Tang, Jin},
  journal={IEEE Transactions on Image Processing},
  volume={30},
  pages={5678--5691},
  year={2021},
  publisher={IEEE}
}

@article{wang2021cgfnet,
  title={CGFNet: Cross-guided fusion network for RGB-T salient object detection},
  author={Wang, Jie and Song, Kechen and Bao, Yanqi and Huang, Liming and Yan, Yunhui},
  journal={IEEE Transactions on Circuits and Systems for Video Technology},
  volume={32},
  number={5},
  pages={2949--2961},
  year={2021},
  publisher={IEEE}
}

@inproceedings{guerrero2015extremely,
  title={Extremely overlapping vehicle counting},
  author={Guerrero-G{\'o}mez-Olmedo, Ricardo and Torre-Jim{\'e}nez, Beatriz and L{\'o}pez-Sastre, Roberto and Maldonado-Basc{\'o}n, Saturnino and Onoro-Rubio, Daniel},
  booktitle={Iberian conference on pattern recognition and image analysis},
  pages={423--431},
  year={2015},
  organization={Springer}
}

@inproceedings{sindagi2019multi,
  title={Multi-level bottom-top and top-bottom feature fusion for crowd counting},
  author={Sindagi, Vishwanath A and Patel, Vishal M},
  booktitle={Proceedings of the IEEE/CVF international conference on computer vision},
  pages={1002--1012},
  year={2019}
}

@inproceedings{lin2022boosting,
  title={Boosting crowd counting via multifaceted attention},
  author={Lin, Hui and Ma, Zhiheng and Ji, Rongrong and Wang, Yaowei and Hong, Xiaopeng},
  booktitle={Proceedings of the IEEE/CVF conference on computer vision and pattern recognition},
  pages={19628--19637},
  year={2022}
}

@inproceedings{ma2021learning,
  title={Learning to count via unbalanced optimal transport},
  author={Ma, Zhiheng and Wei, Xing and Hong, Xiaopeng and Lin, Hui and Qiu, Yunfeng and Gong, Yihong},
  booktitle={Proceedings of the AAAI Conference on Artificial Intelligence},
  volume={35},
  number={3},
  pages={2319--2327},
  year={2021}
}

@inproceedings{ma2020learning,
  title={Learning scales from points: A scale-aware probabilistic model for crowd counting},
  author={Ma, Zhiheng and Wei, Xing and Hong, Xiaopeng and Gong, Yihong},
  booktitle={Proceedings of the 28th ACM International Conference on Multimedia},
  pages={220--228},
  year={2020}
}

@inproceedings{lin2021direct,
  title={Direct Measure Matching for Crowd Counting},
  author={Lin, Hui and Hong, Xiaopeng and Ma, Zhiheng and Wei, Xing and Qiu, Yunfeng and Wang, Yaowei and Gong, Yihong},
  booktitle={Proceedings of the Thirtieth International Joint Conference on Artificial Intelligence (IJCAI-21)},
  year={2021},
  publisher={International Joint Conferences on Artificial Intelligence Organization},
}

@inproceedings{lin2024gramformer,
  title={Gramformer: learning crowd counting via graph-modulated transformer},
  author={Lin, Hui and Ma, Zhiheng and Hong, Xiaopeng and Shangguan, Qinnan and Meng, Deyu},
  booktitle={Proceedings of the AAAI Conference on Artificial Intelligence},
  volume={38},
  number={4},
  pages={3395--3403},
  year={2024}
}

@inproceedings{lian2019density,
  title={Density map regression guided detection network for rgb-d crowd counting and localization},
  author={Lian, Dongze and Li, Jing and Zheng, Jia and Luo, Weixin and Gao, Shenghua},
  booktitle={Proceedings of the IEEE/CVF Conference on Computer Vision and Pattern Recognition},
  pages={1821--1830},
  year={2019}
}

@article{lian2021locating,
  title={Locating and counting heads in crowds with a depth prior},
  author={Lian, Dongze and Chen, Xianing and Li, Jing and Luo, Weixin and Gao, Shenghua},
  journal={IEEE Transactions on Pattern Analysis and Machine Intelligence},
  volume={44},
  number={12},
  pages={9056--9072},
  year={2021},
  publisher={IEEE}
}

@article{lesani2020development,
  title={Development and evaluation of a real-time pedestrian counting system for high-volume conditions based on 2D LiDAR},
  author={Lesani, Asad and Nateghinia, Ehsan and Miranda-Moreno, Luis F},
  journal={Transportation research part C: emerging technologies},
  volume={114},
  pages={20--35},
  year={2020},
  publisher={Elsevier}
}

@inproceedings{chen2016svm,
  title={SVM based people counting method in the corridor scene using a single-layer laser scanner},
  author={Chen, Ziqing and Yuan, Wei and Yang, Ming and Wang, Chunxiang and Wang, Bing},
  booktitle={2016 IEEE 19th International Conference on Intelligent Transportation Systems (ITSC)},
  pages={2632--2637},
  year={2016},
  organization={IEEE}
}

@inproceedings{sindagi2020learning,
  title={Learning to count in the crowd from limited labeled data},
  author={Sindagi, Vishwanath A and Yasarla, Rajeev and Babu, Deepak Sam and Babu, R Venkatesh and Patel, Vishal M},
  booktitle={European Conference on Computer Vision},
  pages={212--229},
  year={2020},
  organization={Springer}
}

@inproceedings{lin2023optimal,
  title={Optimal transport minimization: Crowd localization on density maps for semi-supervised counting},
  author={Lin, Wei and Chan, Antoni B},
  booktitle={Proceedings of the IEEE/CVF Conference on Computer Vision and Pattern Recognition},
  pages={21663--21673},
  year={2023}
}

@inproceedings{meng2021spatial,
  title={Spatial uncertainty-aware semi-supervised crowd counting},
  author={Meng, Yanda and Zhang, Hongrun and Zhao, Yitian and Yang, Xiaoyun and Qian, Xuesheng and Huang, Xiaowei and Zheng, Yalin},
  booktitle={Proceedings of the IEEE/CVF international conference on computer vision},
  pages={15549--15559},
  year={2021}
}

@inproceedings{zhao2020active,
  title={Active crowd counting with limited supervision},
  author={Zhao, Zhen and Shi, Miaojing and Zhao, Xiaoxiao and Li, Li},
  booktitle={European conference on computer vision},
  pages={565--581},
  year={2020},
  organization={Springer}
}

@inproceedings{wu2022multimodal,
  title={Multimodal crowd counting with mutual attention transformers},
  author={Wu, Zhengtao and Liu, Lingbo and Zhang, Yang and Mao, Mingzhi and Lin, Liang and Li, Guanbin},
  booktitle={2022 IEEE International Conference on Multimedia and Expo (ICME)},
  pages={1--6},
  year={2022},
  organization={IEEE}
}

@article{wang2025asymmetric,
  title={Asymmetric Modal Fusion for Multi-modal Crowd Counting},
  author={Wang, Chenhao and Hong, Xiaopeng and Ma, Zhiheng and Wang, Yabin and Wei, Yupeng and Zhang, Jinpeng},
  journal={Pattern Recognition},
  pages={112768},
  year={2025},
  publisher={Elsevier}
}

@inproceedings{arazo2020pseudo,
  title={Pseudo-labeling and confirmation bias in deep semi-supervised learning},
  author={Arazo, Eric and Ortego, Diego and Albert, Paul and O’Connor, Noel E and McGuinness, Kevin},
  booktitle={2020 International joint conference on neural networks (IJCNN)},
  pages={1--8},
  year={2020},
  organization={IEEE}
}

@inproceedings{shang20262d,
  title={2d gaussians spatial transport for point-supervised density regression},
  author={Shang, Miao and Hong, Xiaopeng},
  booktitle={Proceedings of the AAAI Conference on Artificial Intelligence},
  volume={40},
  number={11},
  pages={8824--8832},
  year={2026}
}

\end{document}